\documentclass{article}

\usepackage{microtype}
\usepackage{booktabs} 
\usepackage{subcaption}

\usepackage{hyperref}

\usepackage[preprint]{icml2023}

\usepackage{amsmath}
\usepackage{amssymb}
\usepackage{mathtools}
\usepackage{amsthm}
\usepackage[capitalize,noabbrev]{cleveref}

\usepackage{xcolor, soul}
\sethlcolor{red}

\theoremstyle{plain}

\theoremstyle{definition}

\theoremstyle{remark}

\usepackage[textsize=tiny]{todonotes}

\usepackage{arydshln} 
\usepackage{enumitem} 
\newcommand{\mycomment}[1]{}

\newcommand{\ra}[1]{\renewcommand{\arraystretch}{#1}}

\newcommand{\ctxt}[1]{\multicolumn{2}{@{}p{1.0\columnwidth}}{#1}}
\newcommand{\bb}[1]{\textbf{#1}}

\newcommand{\tjo}[1]{}
\icmltitlerunning{Stable Entropy Hypothesis and Entropy-Aware Decoding}

\begin{document}

\twocolumn[
\icmltitle{
The Stable Entropy Hypothesis and Entropy-Aware Decoding: \\An Analysis and Algorithm for Robust Natural Language Generation
}




\begin{icmlauthorlist}
\icmlauthor{Kushal Arora}{mila_mcgill}
\icmlauthor{Timothy J. O'Donnell}{mila_mcgill}
\icmlauthor{Doina Precup}{mila_mcgill,deepmind}
\icmlauthor{Jason Weston}{meta_ai}
\icmlauthor{Jackie C.K. Cheung}{mila_mcgill}
\end{icmlauthorlist}

\icmlaffiliation{mila_mcgill}{Mila \& McGill University}
\icmlaffiliation{deepmind}{DeepMind}
\icmlaffiliation{meta_ai}{Meta AI}

\icmlcorrespondingauthor{Kushal Arora}{kushal.arora@mail.mcgill.ca}

\icmlkeywords{Machine Learning, ICML, Natural Language Degeneration, Natural Language Generation, NLP, Language Modeling, Decoding Methods}

\vskip 0.3in
]



\printAffiliationsAndNotice{}  

\begin{abstract}
  State-of-the-art language generation models can degenerate when applied to open-ended generation problems such as text completion, story generation, or dialog modeling. This degeneration usually shows up in the form of incoherence, lack of vocabulary diversity, and self-repetition or copying from the context. In this paper, we postulate that ``human-like'' generations usually lie in a narrow and nearly flat entropy band, and violation of these entropy bounds correlates with degenerate behavior. Our experiments show that this stable narrow entropy zone exists across models, tasks, and domains and confirm the hypothesis that violations of this zone correlate with degeneration. We then use this insight to propose an entropy-aware decoding algorithm that respects these entropy bounds resulting in less degenerate, more contextual, and "human-like" language generation in open-ended text generation settings.~\footnote{Implementation of Entropy Aware Decoding: \url{https://github.com/kushalarora/transformers/blob/main/src/transformers/generation_utils.py\#L1894}.\\
  Source code for the stable entropy hypothesis analysis and the entropy-aware decoding experiments: \url{https://github.com/kushalarora/stable_entropy_hypothesis}}
\end{abstract}
\section{Introduction}
Current state-of-the-start transformer-based~\citep{vaswani_attention_2017} large language models 
have made a tremendous amount of progress on both strongly conditioned generation tasks such as summarization~\cite{zhang_etal_2020_pegasus,lewis-etal-2020-bart} and machine translation~\cite{raffel_etal_2020_exploring,liuMultilingualDenoisingPretraining2020a} and more open-ended generation tasks such as dialog generation~\cite{rollerRecipesBuildingOpendomain2020,shusterBlenderBotDeployedConversational2022}, story generation~\cite{brownLanguageModelsAre2020}, etc.
Almost all these large-scale language models are trained by maximizing the log-likelihood of the training sequences. 
This, one would assume, will result in likelihood-maximizing decoding algorithms such as greedy and beam search, producing outputs that match the informativeness, coherence, and quality of generation of the training data. 
This assumption holds for more strongly conditioned tasks but, for more open-ended generation tasks, deterministic decoding methods produce repetitive and dull outputs, referred to as degeneration in ~\citet{holtzmanCuriousCaseNeural2019}.
In these open-ended generation settings, stochastic decoding methods can help. These methods uniformly sample from either an annealed or a truncated distribution and are known to produce more coherent generations with less repetition that score high on generation quality metrics such as Mauve~\cite{pillutlaMAUVEMeasuringGap2021} and human acceptability judgments.



In this paper, we examine this degeneration conundrum  --- i.e.,  the degeneration of deterministic decoding methods in an open-ended generation setting while being robust for strongly conditioned generation tasks, and the relative robustness of well-tuned stochastic decoding methods in open-ended generation setups, 
through the lens of entropy of the conditional distribution of the language model\footnote{For brevity, we will refer to the entropy of the conditional distribution of the model as entropy from hereon.}. We start by presenting a finding that, under the context distribution from the training data, the mean entropy of a language model remains stable over the length of the generation. We refer to this mean entropy as the \textit{stable entropy baseline}, and a narrow band around the stable baseline spanned as the \textit{stable entropy zone}. In our experiments, we establish that the stable entropy phenomenon exists across the tasks, domains, and model combinations.

In our analysis, we observe that, in an open-ended generation setting, deterministic decoding algorithms suffer a catastrophic drop in entropy over the sequence length. In contrast, entropy under well-tuned stochastic decoding algorithms remains mostly confined within the stable entropy zone. We use this finding to posit that any decoding algorithm whose resultant entropy across timesteps stays mostly within this narrow stable entropy zone, will result in more coherent and less degenerate text. We refer to this hypothesis as the \textit{stable entropy hypothesis} (SEH). We empirically validate this hypothesis by showing a strong correlation between generation quality and entropy zone violations.


Next, we attempt to explain the relative robustness of deterministic decoding methods such as beam search for strongly grounded tasks using the stable entropy analysis. The stable entropy hypothesis would suggest that for strongly conditioned tasks, we will not observe a similar catastrophic drop in entropy under greedy decoding. Our experiments confirm this observation; i.e., for strongly conditioned generation tasks, the smoothed entropy under beam search mostly respects the stable entropy zone bounds.


Finally, we leverage the stable entropy analysis to propose a new entropy-aware decoding method that can avoid degeneration while acting greedily most of the time. On two tasks: text completion and dialogue generation, we show that entropy-aware decoding results in a less degenerate, more contextually appropriate, and ``human-like" generation.

Summarizing, in this paper, we analyze the disparate behavior of various decoding algorithms through an entropy-aware perspective. We show that deterministic decoding's catastrophic drop in entropy may explain their degeneracy in open-ended generation settings and that this catastrophic drop is absent in strongly grounded tasks indicating their relative robustness in tasks such as machine translation and summarization. We also show that sampling-based methods in open-ended generation setups do respect the stable entropy hypothesis and this adherence to the stable entropy hypothesis strongly correlates with generation quality. Finally, we leverage the insights from our analysis to propose a novel entropy-aware decoding method that results in a more "human-like" generation while acting greedily most of the time.

\section{Stable Entropy Analysis}
\subsection{Stable Entropy Zone}
\begin{figure}[t!]
  \centering
  \includegraphics[width=\columnwidth]{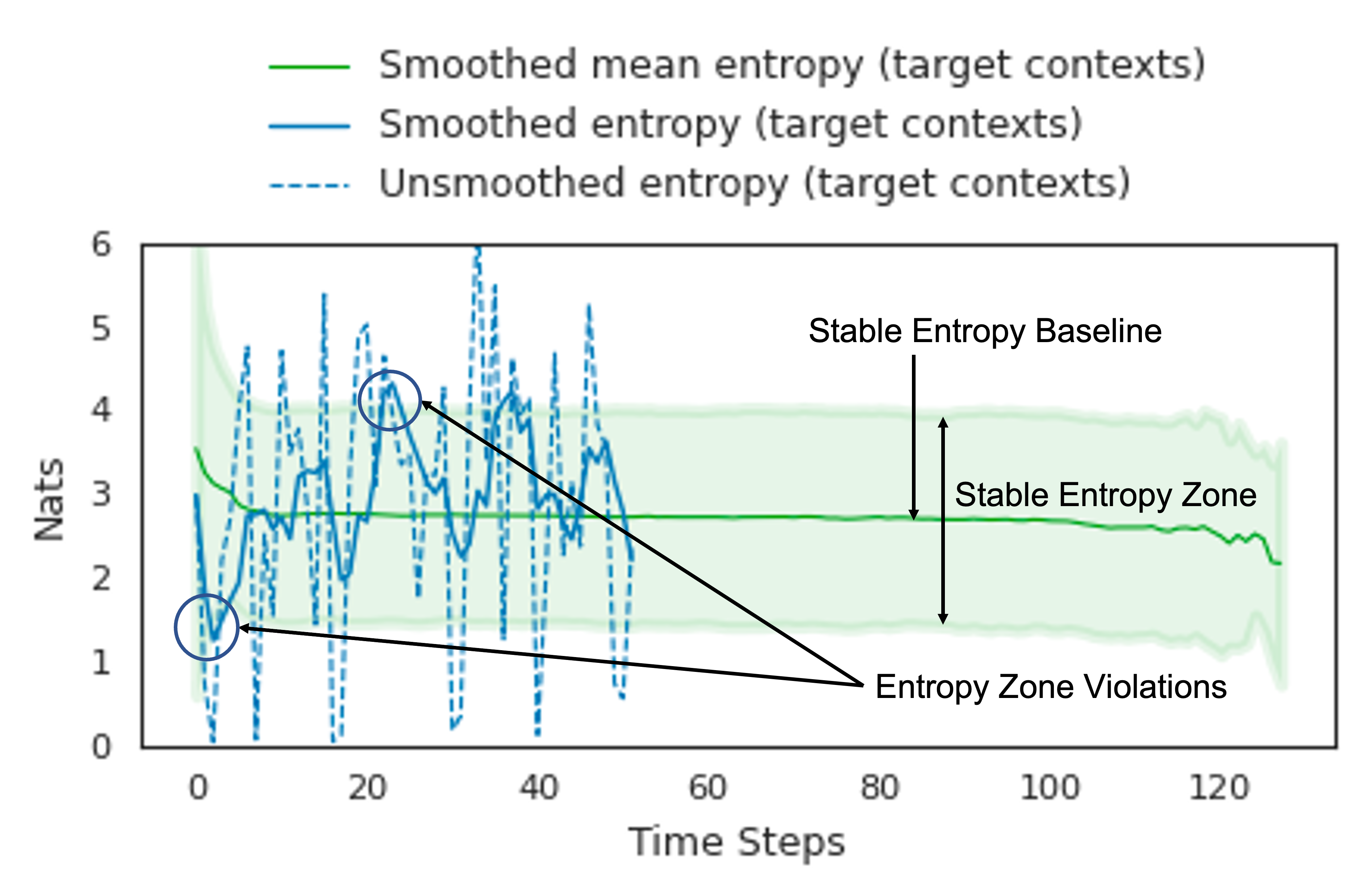}
  \caption{\textbf{The Stable Entropy Zone annotated.} The dashed and solid blue lines represent the entropy and smoothed entropy of single target completion. The faint green line is the mean smoothed entropy computed under the target context distribution. We refer to it as the \textbf{stable entropy baseline}. The green hue around it represents its $1.5$ standard deviation and is the \textbf{stable entropy zone}. Breaches of the stable entropy zone's upper and lower bound are referred to as \textbf{entropy upper bound violations (EUV)} and \textbf{entropy lower bound violations (ELV)} respectively and combined they are called \textbf{entropy bound violations (EV)}.}
  \label{fig:stable_entropy_zone_annotated}
\end{figure}


Let $p_\theta$ be an autoregressive language model trained on a dataset $\mathcal{D}$, parameterized by $\theta$. Given an input or source, $x$, and previously generated tokens or context, $w_{1}^{t}$, the entropy of the model is defined as
\begin{equation}
  \mathcal{H}(p_\theta, w_{1}^{t};x) = \underset{w \sim p_\theta(\cdot| w_{1}^{t})}{\mathbb{E}}  -\log p_\theta(w| w_{1}^{t};x)
\end{equation}
The entropy of the model can suffer from high variance (See Figure~\ref{fig:stable_entropy_zone_annotated}). This variance can possibly be attributed to linguistic and tokenization phenomena such as collocations, the presence of function words, multi-token words, abbreviations, and punctuation in the sequence. To reduce the variance for our analysis, we smooth out the entropy. We compute smoothed entropy at time step $t$ by averaging entropy over a small number ($U$) of previous steps\footnote{We drop the input $x$ from equations for brevity.}:
\begin{equation}
  \bar{\mathcal{H}}(p_\theta, w_{1}^{t}) = 1/U \sum_{j=t-U}^{t} \mathcal{H}(p_\theta, w_{1}^{j}).
\end{equation}
We now define the \textbf{stable entropy baseline} as the mean smoothed entropy at timestep $t$ under the target context distribution at time $t$,  $w_1^t \in \mathcal{D}$:
\begin{equation}
  \mu_{\bar{\mathcal{H}}}(t;\mathcal{D}, p_\theta) = \mathbb{E}_{w_{1}^{t} \in \mathcal{D}} \bigl[  \bar{\mathcal{H}}(p_\theta, w_{1}^{t}) \bigl].
\end{equation}
Next, we define the \textbf{stable entropy zone} as a zone around the stable entropy baseline that covers a major fraction of smoothed entropy (across data points in the corpus) of the model under the target distribution. We define it in terms of the model's standard deviation. We choose $1.5$ standard deviation ($\sigma_{\bar{\mathcal{H}}}(t;\mathcal{D}, p_\theta)$) around the stable entropy baseline as the stable entropy zone for our analysis. This span covers approximately 87\% of smoothed conditional entropies induced under target distribution.\footnote{Other choices of the width of the stable entropy zone are possible. In our experiments, we found that similar choices of the width of stable entropy zone do not impact our conclusions.}

\begin{figure*}[t!]
  \centering
    \includegraphics[width=1.0\textwidth]{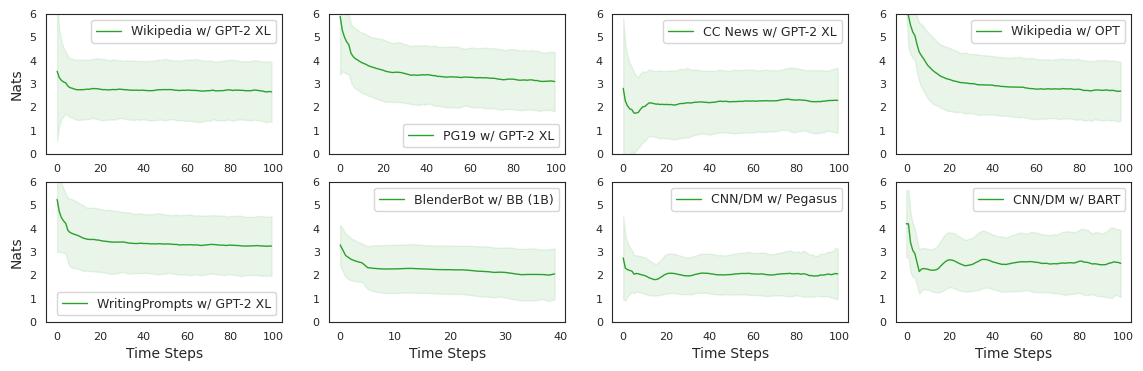}
  \caption{\textbf{Stable entropy baselines across models, tasks, and domains.} We observe that the stable entropy baseline is nearly flat and the stable entropy zone is narrow across models (GPT2-XL, OPT, BlenderBot, Pegasus, and BART), tasks (text completion, story completion, dialog, and summarization), and domains (news, Wikipedia, and fiction).}
  \label{fig:stable_entropy_generalizes}
\end{figure*}
\subsubsection{Empirical Study of Stability}~\label{sec:stable_entropy_zone_experiments}
In this section, we show that the smoothed mean entropy of a model under the target context distribution remains stable; i.e., it remains nearly flat---hence justifying the moniker of the stable entropy baseline. We also show that the stable entropy zone---i.e., the area spanned by a fraction of its standard deviation---is narrow and mostly flat. We start by demonstrating this phenomenon in a text completion setup, then show that the stable entropy zone generalizes across models, domains, and tasks.
\paragraph{Models and Data}
For our text completion experiments, we use the GPT-2 XL~\cite{radford2019language} model and Wikipedia data. We follow a similar setup as \citet{krishnaRankGenImprovingText2022}; i.e., we chunk Wikipedia documents into individual paragraphs and use the first $256$ tokens as prefixes, and limit the generation length to $128$ tokens. 

To demonstrate the generalizability of the stable entropy zone, we use a combination of five tasks, spanning six different datasets and five different models. These tasks are text completion, dialog generation, summarization, and story generation.
 For text completion analysis, we use two models, GPT2-XL~\cite{brownLanguageModelsAre2020} and OPT (1.3B)~\cite{zhangOPTOpenPretrained2022} and three different datasets from three different domains: the Wikipedia dataset~\cite{krishnaRankGenImprovingText2022}, a fiction dataset,  PG19~\cite{rae-compressive-2019}, and a news dataset, CC News~\cite{Hamborg2017}. We evaluate CNN-DM~\cite{hermannTeachingMachinesRead2015} dataset with the BART~\cite{lewis-etal-2020-bart} and the Pegasus~\cite{zhang_etal_2020_pegasus} models for summarization experiments and the BlenderBot (1B)~\cite{rollerRecipesBuildingOpendomain2020} model on the Blended Skill Talk~\cite{smith-etal-2020-put} for dialog generation experiments. For story generation, we evaluate the WritingPrompts~\cite{fanHierarchicalNeuralStory2018} dataset with the GPT-2 XL~\cite{brownLanguageModelsAre2020} model.

 \subsubsection{Results}
\paragraph{Stable Entropy Zone Does Exist.}
\autoref{fig:stable_entropy_zone_annotated} shows the stable entropy baseline and the stable entropy zone computed in the Wikipedia text completion setting. The blue dotted line shows the entropy of the model under the target context distribution for a single prefix (See Appendix Table~\ref{tab:decoding_generations} for the prefix and the target completion). We can observe in \autoref{fig:stable_entropy_zone_annotated} that the unsmoothed entropy of the model contains many sudden drops or peaks thus necessitating the need to smooth it out for analysis. The solid blue line represents this smoothed entropy under the target context distribution. The solid green line shows the stable entropy baseline. As the figure shows, the mean smoothed entropy under the target context distribution (i.e., the stable entropy baseline) remains nearly flat except for the first few steps. The region around the stable entropy baseline represented with a green hue is the stable entropy zone. We can observe that the stable entropy zone is narrow ($\approx 2$ Nats) and flat.
\begin{table}[t!]
  \small
  \centering \ra{1.0} \scalebox{1}{
  \begin{tabular}{@{}lccc@{}}
  \toprule
   \textbf{Setup}  &  \textbf{MSE} &  \textbf{Slope} &  \textbf{Intercept} \\ 
  \midrule
  GPT2-XL/Wiki (TC) & 0.010 & -0.0028 & 2.88 \\
  GPT2-XL/PG19 (TC) & 0.090 & -0.009 & 3.91 \\
  OPT/Wiki (TC)     &0.207 & -0.0130 & 3.80 \\
  BB1/BST (D)       &0.011 & -0.0143 & 2.64 \\
  BART/CNN-DM (S)   & 0.069 & 0.0071 & 2.74 \\
  \bottomrule
  \end{tabular}}
  \caption{\textbf{Can stable entropy baseline be modeled using a flat 1-D line?}. Modeling smoothed mean entropy using a 1-D line yields a very small mean squared error and a near-zero slope. This indicates a flat 1-D line can model the stable entropy baseline.}
  \label{tab:stable_entropy_losses}
\end{table}
\paragraph{Stable Entropy Zone Generalizes Across Tasks, Domains, and Models.}
\autoref{fig:stable_entropy_generalizes} shows the stable entropy baselines and the stable entropy zones across a combination of different tasks, models, and domains. Again, we observe that, except for the first few steps, the stable entropy baseline remains almost always flat and that the stable entropy zone almost always forms a narrow and flat band around it. 

We further quantify this observation by finding the line of best fit for the smoothed mean entropy (Table~\ref{tab:stable_entropy_losses}). We observe very low mean squared error loss and near-zero slope coefficients indicating that a flat 1-D line can fit smoothed entropy under the target context distribution with very few outliers. The variance in intercepts indicates that the stable entropy baseline must be evaluated individually for each model and dataset combination.
\begin{figure*}[t!]
  \centering
    \includegraphics[width=0.9\textwidth]{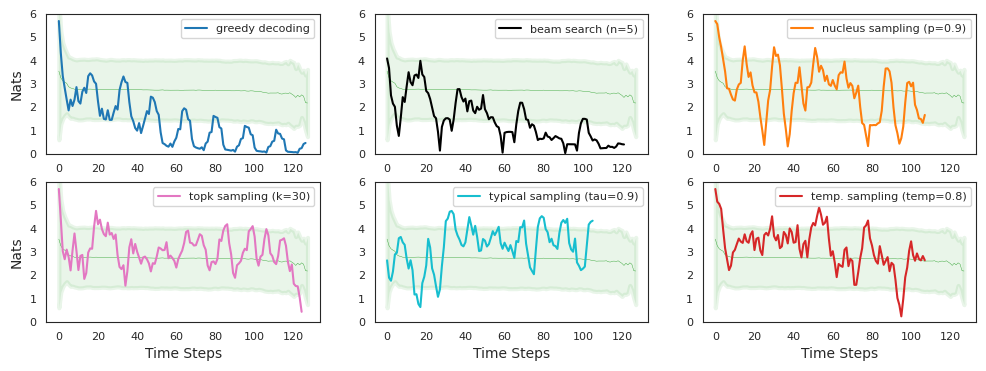}
  \caption{\textbf{Visualization of entropy of various decoding algorithms.} Visualizing the smoothed entropy for various decoding algorithms in a text completion setup given a prompt. We observe the catastrophic entropy drop in the case of the beam and greedy search. Stochastic algorithms try to stay in the stable entropy zone. \autoref{tab:decoding_generations} shows the prompt and generations corresponding to these visualizations.}
  \label{fig:decoding_algorithm_entropy_visualizations}
\end{figure*}
\subsection{Stable Entropy Hypothesis}
\autoref{fig:decoding_algorithm_entropy_visualizations} visualizes the completions for a given single prefix from Appendix Table~\ref{tab:decoding_generations} for various decoding algorithms in the Wikipedia text completion.
We can clearly observe a catastrophic drop in smoothed entropy for beam and greedy search whereas smoothed entropy of well-tuned sampling-based decoding algorithms stays mostly within the stable entropy zone. These stochastic decoding algorithms are also known to produce better completions \cite{holtermann-etal-2022-fair}. 
We postulate that these two things might be related. 

We hypothesize that decoding algorithms whose generation's smoothed entropy stays mostly enclosed within the stable entropy zone will produce higher quality, coherent, less repetitive, and more ``human-like" text.  We refer to this hypothesis as the \textbf{stable entropy hypothesis}. 

Next, we empirically verify the stable entropy hypothesis by measuring the correlation between automatic metrics of text generation quality and entropy violation measures.

\subsubsection{Experiments}~\label{sec:stable_entropy_experiments}
In this section, we answer the following two questions: 
\setlist{nolistsep}
\begin{itemize}[noitemsep]
  \item Are violations of the stable entropy zone correlated with automatic measures of generation quality in more open-ended generation settings?
  \item Does the stable entropy hypothesis also hold for more strongly conditioned tasks where deterministic search strategies do not degenerate?
\end{itemize}
\vspace{-1.0em}
\begin{figure*}[t!]
  \centering
  \scalebox{1}[1]{
  \begin{subfigure}{0.33\textwidth}
    \centering
    \includegraphics[width=1\textwidth]{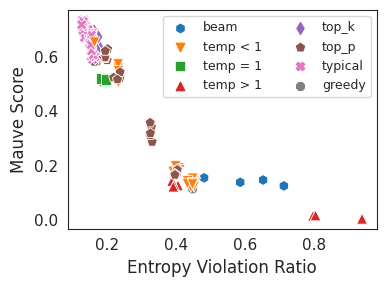}
    \caption{EV Ratio vs Mauve Score}
    \label{fig:evr_vs_mauve}
  \end{subfigure}
  \begin{subfigure}{0.33\textwidth}
    \centering
    \includegraphics[width=1\textwidth]{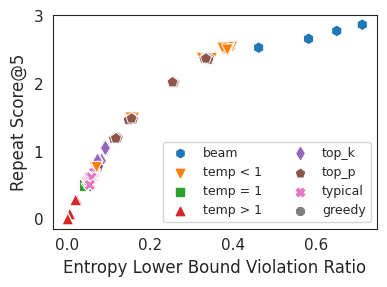}
    \caption{ELV Ratio vs Repeat Score@5}
    \label{fig:elvr_vs_repeat}
  \end{subfigure}
  \begin{subfigure}{0.33\textwidth}
    \centering
    \includegraphics[width=1\textwidth]{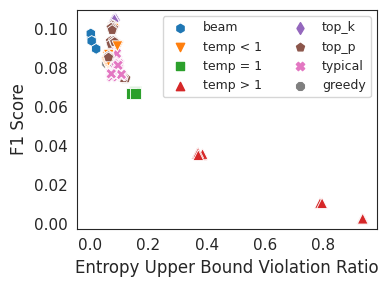}
    \caption{EUV Ratio vs F1 Score}
    \label{fig:f1_score_vs_euvr}
  \end{subfigure}}
  \caption{\textbf{Entropy violations vs repetition vs generation quality vs coherence.} Figure (\protect\subref{fig:evr_vs_mauve}) shows that the Mauve score, a proxy for generation quality, correlates negatively ($\rho=-0.92$) with the entropy violations. Figure (\protect\subref{fig:elvr_vs_repeat}) shows lower entropy violations are strongly correlated ($\rho=0.96$) with the repetition issue. Finally, Figure~(\protect\subref{fig:f1_score_vs_euvr}) shows that decodings schemes that result in high entropy produce relatively more incoherent text ($\rho=-0.93$).}
  \label{fig:entropy_vs_mauve_vs_repeat}
\end{figure*}
\paragraph{Models, Data, and Metrics:}
We answer the first question in the same text completion setting as discussed in Section~\ref{sec:stable_entropy_zone_experiments}; i.e., we use the GPT-2 XL~\cite{radford2019language} model and Wikipedia data from \citet{krishnaRankGenImprovingText2022}. In this setting, we evaluate various configurations of well-known decoding algorithms, namely, top-k sampling~\cite{holtzmanCuriousCaseNeural2019}, nucleus sampling~\cite{fanHierarchicalNeuralStory2018}, temperature sampling, and typical decoding~\cite{meister_2023_locally_typical}. See Appendix Section~\ref{sec:decoding_configs} for the configurations.

We use four automatic metrics to evaluate the performance of various decoding algorithms. \textbf{F1} computes the overlap between the generation and the ``true" completion of the prefix, indicating whether the text is on-topic and contextually appropriate.\footnote{We filter out stop words from the sequences before computing F1 scores to ensure that these commonly occurring words do not confound contextuality judgment.} \textbf{Repeat Score@5} cumulatively measures the repetition across 1- to 5-grams weighted exponentially and normalized by length.~\footnote{We discuss the exact computation of Repeat Score@5 in the Appendix Section~\ref{sec:repeat_score_5_def}.} \textbf{3-gram repeats} measures the number of 3-gram repeats in the generated sequence. A higher Repeat Score@5 and 3-gram repeats indicate that the generation was more repetitive and dull. \textbf{Mauve}~\cite{pillutlaMAUVEMeasuringGap2021}, a recently introduced automatic generation quality metric, evaluates generation quality in the open-ended generation setting and was shown to have a strong correlation with human acceptability judgments. 

We measure entropy zone violations using three metrics. \textbf{entropy lower-bound violation ratio (ELVR)} measures the ratio of instances when smoothed entropy falls below the lower bound of the stable entropy zone. Similarly,  \textbf{entropy upper-bound violation ratio (EUVR)} measures the ratio of instances where smoothed entropy is larger than the upper bound of the stable entropy zone. The third metric, \textbf{entropy violation ratio (EVR)}, is the sum of the two ratios and measures the ratio of instances when entropy falls outside either the lower or the upper bound.

To answer the second question, we contrast the performance of beam search with a fixed beam size ($n=5$) between strongly conditioned tasks and more open-ended generation tasks. We use summarization and machine translation as our prototypical strongly conditioned language generation tasks and text completion and dialog generation as open-ended generation tasks. For summarization experiments, we report the ROUGE-1 score on two datasets and model combinations. We evaluate the CNN-DM dataset with the Pegasus~\cite{zhang_etal_2020_pegasus} model and the Arxiv~\cite{cohan-etal-2018-discourse} dataset with the BigBird-Pegasus~\cite{zaheer_big_bird_2020} model. We report BLEU scores for our machine translation experiment on the WMT 2017 dataset (de-en split)~\cite{bojar-EtAl:2017:WMT1} with two models, 
M-BART~\cite{tang-etal-2021-multilingual} and Opus~\cite{tiedemann-thottingal-2020-opus}. We contrast the performance of beam search on these tasks with two more open-ended generation tasks namely, text completion and dialog generation. We report the F1 score for dialog generation on the Blended Skills Talk dataset~\cite{smith-etal-2020-put} with two different sizes of the BlenderBot model~\cite{rollerRecipesBuildingOpendomain2020} (90M and 1B parameters). For text completion experiments, we report results on the Wikipedia dataset from \citet{krishnaRankGenImprovingText2022} with GPT2-XL~\cite{radford2019language} and OPT (1.3B)~\cite{zhangOPTOpenPretrained2022} models. Except for text completion models, all the other models are encoder-decoder models.

\subsubsection{Results}
\paragraph{Stable Entropy Hypothesis Holds for Text Completion.} 

We present the correlation results in the text completion setting in \autoref{fig:entropy_vs_mauve_vs_repeat}. We observe that Mauve scores have a strong negative correlation ($\rho=-0.92$) with the entropy violation ratio (EVR). This indicates a decoding algorithm that generates more instances of smoothed entropy falling outside the stable entropy zone usually has worse generation quality. We also observe a strong positive correlation ($\rho=0.96$) between the Repeat Score@5 and the entropy lower-bound violation ratio (ELVR). This matches our observation that deterministic decoding methods which are prone to repetition and copying exhibit a catastrophic drop in smoothed entropy resulting in them falling below the stable entropy zone's lower bound. Finally, we observe a negative correlation ($\rho=-0.93$) between the entropy upper-bound violation ratio and F1 scores, indicating decoding methods with high entropy (e.g., sampling with $t=1.5$) usually produce a less coherent text. We present additional correlation plots in Appendix Section~\ref{sec:additional_corr_plots}.


Appendix Table \ref{tab:stable_entropy_quantitative_results} quantitatively verifies this hypothesis by showing that generations under greedy decoding and beam search degenerate as indicated by low Mauve score and high Repeat Score@5 and 3-gram repeats. This degeneration correlates with a high overall Entropy Violation Ratio (EVR), a significant portion of which are entropy lower bound violations. High entropy upper bound violations, as is the case with sampling with a higher temperature hyperparameter ($t=1.2$), indicate incoherence that can be attributed to a high amount of randomness, as suggested by very low Mauve and F1 scores. And, fewer entropy violations (both upper and lower bound), as in the case of top-k, nucleus, and typical sampling, do correlate in fewer repetitions, reasonable F1 score, and a high Mauve score, suggesting better generation quality under these decoding schemes.

\autoref{fig:decoding_algorithm_entropy_visualizations} visualizes the completions from various decoding algorithms for a single prefix.   We can observe that well-tuned sampling-based decoding algorithms mostly stay enclosed within the stable entropy zone. The prefix and completions used to generate these visualizations are presented in Appendix~\autoref{tab:decoding_generations} and they qualitatively show that the generations produced under sampling-based methods do indeed appear more coherent and less repetitive.

\begin{table*}[h!]
  \small
  \centering \ra{1.3} \scalebox{1}{
  \begin{tabular}{@{}lcccccc@{}}
  \toprule
\textbf{Setup}        & \textbf{Task} & \textbf{Mauve} ($\uparrow$)  & \textbf{F1} ($\uparrow$) & \textbf{BLEU} ($\uparrow$) & \textbf{ROUGE-1} ($\uparrow$) & \textbf{ELVR} ($\downarrow$) \\ 
  \midrule
  GPT2-XL/Wiki      & Text Completion & 0.137 & & & & 0.588 \\
  GPT2-XL/PG19      & Text Completion & 0.048 & & & & 0.690 \\
  OPT/Wiki          & Text Completion & 0.141 & & & & 0.605 \\
  \midrule
  BB-90M/BST        & Dialog & & 0.110 & & & 0.300 \\
  BB-1B/BST         & Dialog & & 0.153 & & & 0.185 \\
  Pegasus/CNN-DM    & Summarization & &  & & 43.37 (-4.6) & 0.121 \\
  BigBird/Arxiv     & Summarization & &  & & 46.05 (-4.9) & 0.147 \\
  Opus/WMT          & Machine Translation & &  & 35.04 (-6.0) & & 0.084 \\
  MBART/WMT         & Machine Translation & &  & 37.90 (-3.2) & & 0.101 \\
  \bottomrule
  \end{tabular}}
  \caption{\textbf{Beam Search does not degenerate for strongly conditioned generation tasks.}. We observe that for conditional generation tasks such as summarization, and machine translation, beam search performs well as indicated by high ROUGE-1 and BLEU scores respectively. The number in the bracket indicates the drop in performance compared to the state of the art.  These tasks and model combinations also have few entropy zone violations. For open-ended tasks such as text competition, the beam search catastrophically degenerates resulting in a poor Mauve score and high entropy violation ratio. See Appendix~\ref{sec:model_and_hyperparams} for model details and state-of-the-art baselines.}
  \label{tab:beam_search_entropy_violations}
\end{table*}
\paragraph{When Beam Search Does Not Degenerate.}
\autoref{tab:beam_search_entropy_violations} presents the generation results decoded using beam search ($n=5$) for both more open-ended (rows 1-5) and strongly conditioned generation settings (rows 6-10). For text completion tasks, we observe a catastrophic drop in entropy and poor generation quality as indicated by high entropy lower-bound violation ratio (ELVR) and low Mauve scores respectively. In contrast, for the strongly-conditioned generation tasks (last 5 rows), the beam search performs substantially better as indicated by near state-of-the-art ROUGE-1 and BLEU scores for summarization and machine translation respectively. Also, we observe far fewer entropy violations for these strongly-conditioned tasks indicating that they mostly remain within the stable entropy zone, thus respecting the stable entropy hypothesis. Dialog generation, an open-ended generation task modeled using an encoder-decoder model falls squarely between the two and has relatively high ELVR scores. A larger and more performant (in terms of F1 score) dialog model (BB-1B) does produce fewer entropy violations.

\section{Entropy Aware Decoding}
\begin{algorithm}[tb!] 
  \caption{Entropy-Aware Decoding}
  \label{algo:entropy_aware_decoding_in_brief}
  \begin{algorithmic}
      \STATE {\bfseries Input:} input $x$, model $p_{\theta}$
      \STATE {\bfseries HP:} sampling $\mathcal{S}$, patience $N$, margin $\alpha$, ngreedy $g$
      \STATE Initialize $n \gets 0$
      \WHILE {$t \le T$}
        \STATE $p_t = p_{\theta}(y_{t-1},x)$
        \STATE $w_t = \text{argmax}(p_t)$ 
        \IF {$t \le g$}
          \STATE continue
        \ENDIF
        \STATE $\mathcal{H}_{t} = \text{Entropy}(p_t)$  
        \IF {\text{AboveStableEntropyZone}($\mathcal{H}_{t}$, $\alpha$)}
          \STATE $w_t = \text{Sample}(p_t, \mathcal{S})$ 
        \ENDIF 
        \IF {\text{BelowStableEntropyZone}($\mathcal{H}_{t}$, $\alpha$)}
            \STATE $n = n + 1$
        \ELSE
          \STATE $n \gets 0$
        \ENDIF 
        \IF {$n > N$}
          \STATE $y_{t-1}, p_t = \text{BackOffTo}(t-N)$ 
          \STATE $t = t - N$ 
          \STATE $w_t = \text{NextRankedToken}(p_{t})$ 
          \STATE $n \gets 0$
          \ENDIF
          \STATE $y_{t-1} = y_{t-1}w_t$
      \ENDWHILE
  \end{algorithmic}
\end{algorithm}

In the previous section, we discussed how well-tuned stochastic decoding methods can alleviate degeneration issues in open-ended generation settings and how this improvement in generation quality also correlates with lower stable entropy zone violations.
These stochastic methods, though, rely on uniform random sampling at each time step, which might results in generation being less contextual and more factually inaccurate~\cite{leeFactualityEnhancedLanguage2022}.
 In this section, we use the insights from the stable entropy analysis to propose a decoding algorithm that can overcome the degeneration issues of deterministic algorithms while still acting greedily most of the time thus avoiding uniform randomness introduced by the stochastic decoding algorithms. The goal is that such an algorithm would result in a coherent and more on-topic and contextually appropriate generation.

The proposed entropy-aware decoding (EAD) method is outlined in Algorithm~\ref{algo:entropy_aware_decoding_in_brief}. Entropy-aware decoding lets the model decode greedily most of the time and intervenes in two scenarios: 1.) when the entropy of the model breaches the upper bound, and 2.) when the entropy of the model breaches the lower bound $N$ consecutive times. We refer to these interventions as \textbf{Entropy Upper-Bound Interventions (EUI)} and \textbf{Entropy Lower-Bound Interventions (ELI)} respectively.

Upper-bound violation of the stable entropy zone indicates that the model is less certain about its prediction. In such scenarios, chances of miscalibration are high; i.e., the most probable token might not be the ``correct" token. Hence, in the scenario where entropy breaches the upper bound of the stable entropy zone we resort to sampling from the conditional distribution. While sampling, we can rely on any of the off-the-shelf methods (denoted by $\mathcal{S}$ in Algorithm~\ref{algo:entropy_aware_decoding_in_brief}) such as top-k, top-p, or typical sampling.

For lower-bound violations, we wait until $N$ consecutive violations, as entropy drop at any time-step might be due to the presence of multi-token words, abbreviations, or other tokenization quirks. If the entropy is below the lower bound of the entropy zone for $N$ consecutive steps, we back off $N$ steps to the index when it was last above the threshold. At that index, we ignore the current most likely token and select the next highest-ranked token. We continue executing the backoff strategy until we select a token that does not lead $N$ consecutive steps of entropy lower-bound violations.

\subsection{Experiments}
\subsubsection{Model and Data}
We benchmark entropy-aware decoding on two open-ended generation tasks: text completion and dialog generation.
\vspace{-0.5em}
\paragraph{Text Completion}
We use a similar setup and metrics as Section~\ref{sec:stable_entropy_experiments} for our text completion experiments. Additionally, we also report \textbf{\%Det}, the percentage of the time entropy-aware decoding and other algorithms act deterministically and \textbf{\#Backoff}, an average number of times entropy-aware decoding algorithm resorted to backoffs.
\vspace{-0.5em}
\paragraph{Dialog Generation}
For dialog generation experiments, we use the 90M parameter BlenderBot model~\cite{rollerRecipesBuildingOpendomain2020} and report results on the Blended Skills Talk dataset~\cite{smith-etal-2020-put}. We flatten the dialogs in the dataset by concatenating the previous utterances in a dialog. This forms the context for generating the next utterance. We limit the size of the context to 80 words and only keep the latest utterances that fit within the context size. We limit the maximum length of a generated utterance to 128 tokens. We report our dialog generation results on three metrics, namely, F1, Repeat Score@5, and entropy violation ratio. For these experiments, we follow the standard practice and do not remove the stop words from the target and the generated utterances while computing the F1 score. 
\subsubsection{Results}
\begin{table*}[h!]
  \small
  \centering  \ra{1.4}\scalebox{1}{
  \begin{tabular}{@{}p{0.28\textwidth}ccccccc@{}}\toprule
  \textbf{Decoding Method} & \textbf{F1} & \textbf{Rep. Score@5} &  \textbf{Mauve} & \textbf{EVR} & \textbf{Det\%} & \textbf{\#Backoffs}\\
  \midrule
  Greedy        &0.082 & 2.542 &  0.114 & 0.447 & 100 & -\\
  \midrule
  Beam (n=5)    & 0.094 & 2.664 &  0.138 & 0.585 & 100 & -\\
  \quad +3-gram block & 0.102 & 0.666  &  0.476 & 0.170 & 100 & -\\
  \midrule
  Typical Sampling ($\tau=0.2$) & 0.076 & 0.507 &  0.697 & 0.129 & 0 & - \\
  \midrule
  Top-k ($k=30$)                & 0.094 & 0.709 & 0.665 & 0.148 & 0 & - \\
   \midrule
  \multicolumn{7}{@{}l}{Entropy-Aware Decoding (ours)} \\
  \midrule
   $\tau=0.2$, $N=5$, $\alpha=0.5$, $g=10$ &0.090 & 0.696  & 0.688 & 0.116 & 58.8 & 1.97 \\
   \quad \text{w/o ELI} & 0.092 & 0.773 & 0.652 & 0.155 & 58.7 & 0 \\
   \hdashline 
   $k=30$, $N=5$, $\alpha=0.5$, $g=5$ &0.100 & 0.941  & 0.683 & 0.135 & 59.6 & 2.88 \\
   \quad \text{w/o ELI} & 0.101 & 1.06 & 0.657 & 0.178 & 59.45 & 0 \\
   \hdashline
   \text{EAD w/o EUI} $N=5$, $\alpha=0.5$ & 0.089 & 2.124 & 0.232 & 0.308 & 100 & 8.90 \\
   \midrule
    Target completions & 1.000 & 0.605  & 1.000 & 0.136 & - & -\\
  \bottomrule
  \end{tabular}}
  
  \caption{\textbf{Entropy-Aware Decoding Text Completion Experiment.} We observe that entropy-aware decoding is competitive with typical sampling, the best performing stochastic decoding method from Table~\ref{tab:stable_entropy_quantitative_results}, on generation quality and repetitions while having higher F1 score indicating more contextually appropriate completions.}
  \label{tab:ead_gpt2_results}
\end{table*}

\begin{table}[t!]
  \small
  \centering  \ra{1.4} \scalebox{1}{
  \begin{tabular}{@{}p{0.14\textwidth}cccc@{}}\toprule
  Decoding Method &F1 & Rep. Score@5 & EVR & Det\% \\
  \midrule
  Greedy              & 0.115 & 1.229 & 0.162 & 100\\
  Beam (n=5)          & 0.118 & 1.171 &  0.305 & 100\\
  \midrule
  Top-k (k=30)        & 0.112 & 0.489 &  0.155 & 0\\
  Nucleus ($p=0.9$)  & 0.116 & \textbf{0.526}  & 0.160  & 0\\
  \midrule
  EAD ($k=30$)      & 0.125 & 0.674 & 0.130 & 62\\
  \quad \text{w/o ELI} & \textbf{0.125} & 0.731 & 0.155 & 64 \\
  \hdashline
  EAD ($p=0.9$) & \textbf{0.125} & 0.685 & \textbf{0.130} & 62 \\
  \quad \text{w/o EUI} & \textbf{0.126} & 0.742 & 0.156 & 64 \\
  \hdashline
  EAD w/o EUI & 0.114 & 1.166 &  0.136 & 100\\
  \bottomrule
  \end{tabular}}
  \caption{\textbf{Entropy-Aware Decoding Dialog Generation Experiments.} We observe that entropy-aware decoding produces the highest F1 score among all the methods irrespective of the choice of sampling algorithm. It achieves this while reducing the repetitions encountered when generating with greedy or beam search.}
  \label{tab:stable_entropy_dialog_results}
\end{table}
\paragraph{Text Completion Results:} 
\autoref{tab:ead_gpt2_results} presents the results for text completion experiments. We can observe that the entropy-aware decoding (with patience window, $N=5$, margin $\alpha=0.8$, and typical sampling with $\tau=0.2$) generates more on-topic and contextually appropriate, less repetitive, and higher quality text as indicated by high F1 score, low Repeat Score@5 and 3-gram repeats, and high Mauve score respectively. Also, the entropy-aware decoding method has the lowest entropy violation ratio supporting our hypothesis that this improved generation quality might be due to entropy-aware decoding's ability to stay within the stable entropy zone. Entropy-aware decoding acts greedily most of the time (nearly 60\%) as indicated by Det\% measure.
\paragraph{Dialog Generation Results:}
Table~\ref{tab:stable_entropy_dialog_results} presents the result for our dialog generation experiments. We observe stochastic decoding methods do reduce repetition but at the cost of a lower F1 score. This reduction in the F1 score can be attributed to uniform randomness introduced by stochastic decoding methods. Entropy-Aware Decoding (patience window, $N=5$, margin $\alpha=0.25$), with both top-k and top-p sampling, successfully reduces the repetition issue while achieving the highest F1 score. 
\paragraph{Human Evaluation} 
We also performed human evaluation comparing $100$ samples the entropy-aware decoding with typical sampling ($\tau=0.2$) to typical sampling ($\tau=0.2$) and $100$ samples of entropy-aware decoding with top-k sampling (k=30) with top-k sampling (k=30). We asked 8 human evaluators to compare 25 samples each, collecting two annotations per sample in the process. EAD ($\tau=0.2$) was preferred over typical sampling $57\%$ of the time whereas EAD ($k=30$) was preferred over top-k sampling $64\%$ of the time.

\section{Discussion and Related Work}

\paragraph{Entropy-based Becoding Approaches:}Recently, a few stochastic methods have been proposed that use entropy or related concepts to truncate the probability distribution.  \textbf{Typical decoding}~\cite{meister_2023_locally_typical} induces sparsity by selecting a subset of tokens whose likelihood is closest to the entropy of the model. The number of tokens is controlled by the cumulative probability we want to retain in the distribution. \textbf{Mirostat decoding}~\cite{basu-mirostat-2020} modifies top-k sampling where the $k$ is dynamic and controlled in such a way that it ensures that the generation has similar perplexity to the target data. Recently proposed \textbf{$\eta$-sampling}~\cite{hewitt-truncation-2022} samples from the tokens whose probability is greater than $\eta$ which is defined as a function of the model's entropy.
All these decoding methods are fully stochastic, sampling at each time step, introducing uniform randomness which might hurt the contextuality and the factuality of the generation~\cite{leeFactualityEnhancedLanguage2022}. In contrast, entropy-aware decoding only samples if the upper bound of the stable entropy zone is violated. This behavior results in a higher F1 score indicating more on-topic and contextual generations.

\paragraph{Stable Entropy Hypothesis and Uniform Information Density Hypothesis:}
Uniform information density (UID) hypothesis~\cite{levy2005probabilistic,jaegerSpeakersOptimizeInformation2006} states that subject to the grammar constraint, humans prefer sentences that distribute information, measured in terms of surprisal, equally across the linguistic signal~\cite{meisterIfBeamSearch2020}. 

The UID hypothesis is related to the stable entropy hypothesis as both predict the "stable" behavior of the model's prediction under human context distribution. But they also differ in some crucial aspects. First, the stable entropy hypothesis is defined in the terms of entropy, which is expected surprisal over vocabulary under the model distribution. Second, the stable entropy hypothesis is more accommodating as it just expects the model's entropy to fall within a narrow zone whereas the UID hypothesis expects the model's generation's surprisal to be nearly flat or stable for it to ``human-like''. We plot surprisal for the same prefix as in \autoref{fig:decoding_algorithm_entropy_visualizations} in Appendix \autoref{fig:decoding_algorithm_surprisal_visualizations}. Similar to the catastrophic drop in entropy under greedy and beam search, we observe that greedy and beam search do not follow the UID hypothesis and suffer a similar drop in surprisal.


\paragraph{Stable Entropy Hypothesis and Expected Information Hypothesis:}
The Expected Information Hypothesis, proposed by \cite{meister-etal-2022-high}, formally states that text perceived as human-like typically encodes an amount of information close to the expected information content of natural language strings i.e., in the interval $[H(p) - \epsilon, H(p) + \epsilon]$ for a natural language. Text that falls outside of this region is likely perceived as unnatural.

Though both hypotheses state that a ``Goldilocks'' zone exists around an anchor entropy value(s) for natural-sounding language generation, they differ on key operational details. First, the anchor entropy value in the case of the expected information hypothesis is computed under ancestral sampling. The choice of ancestral sampling (temperature sampling with $temp=1$) for defining the anchoring entropy value is questionable given that the decoding approach itself performs relatively poorly on text quality (low Mauve score) and contextuality (low F1 score) metrics.


Additionally, the stable entropy hypothesis has an inherent temporal element to it which is missing in the expected information hypothesis. This can be useful to analyze evolving behavior of various decoding algorithms over the course of the generation such as a correlation between repetitions exhibited by the greedy and beam search and the catastrophic drop in entropy, or the ancestral sampling with high temperature (temp=1.5) and steadily increasing entropy.

\paragraph{Stable Entropy Hypothesis and Local Typicality:} 
~\citet{meister_2023_locally_typical} define the concept of local typicality and hypothesize that natural-sounding language belongs to the local typical set for the human language process. The authors further use this concept to propose typical sampling, one of the decoding algorithms evaluated in the paper. 

~\citet{meister_2023_locally_typical} define $(T,\epsilon)$\textbf{-locally typical set} of language process $\textbf{Y}$ ($\textbf{Y}=\{Y_t\}_{t=1}^{\infty}$) under distribution $p$, as a set of all sequences of length $T$ such that 
\begin{multline}
  \mathcal{L}_{\epsilon}^{(T)} = \big\{\textbf{y}=y_{0}\cdots y_{T}| \forall 1 \le t \le T, \\
      \big| \log p(y_t|\textbf{y}_{<t}) + H(Y_t| \textbf{Y}_{<t}=\textbf{y}_{<t}) \big| < \epsilon \big\}
\end{multline}
In words, this means that we should expect every word in natural-sounding sentences to be close to the expected information content under $p$, i.e., the conditional entropy given prior context.~\cite{meister_2023_locally_typical}.

This notion of local typicality differs from the stable entropy hypothesis in two crucial ways. First, local typicality bounds the surprisal or the information content measured using its negative log probability. The stable entropy hypothesis, in contrast, bounds the entropy of the conditional distribution of the model~\footnote{\citet{meister_2023_locally_typical} refer to this as conditional entropy.}. Second, the stable entropy zone is anchored around the stable entropy baseline which is defined in terms of the entropy of the model under target distribution whereas local typicality uses the entropy of the model under the distribution induced by the current decoding algorithm. Thus, this definition cannot be used to analyze the decoding algorithms' behaviors. A case in point is the analysis of degenerate behavior under deterministic decoding in an open-ended generation setting. In this setting, the anchor value---i.e., the entropy of the model under greedy decoding, will itself drop catastrophically resulting in surprisal always staying within the bounds, indicating that strings generated under greedy decoding satisfy local typicality and hence are natural sounding. This conclusion does not hold as generated text almost always degenerates under greedy decoding in open-ended generation tasks.

\section{Conclusion}
In this paper, we presented the stable entropy hypothesis which states that the entropy of natural language stays in a narrow zone around the stable baseline which is defined as the mean entropy of the model under the target context distribution. We verify this hypothesis in the text completion setting, showing that fewer violations of the stable entropy zone correlate with fewer repetitions and higher generation quality. Next, we leveraged this analysis to propose a mostly-deterministic entropy-aware decoding method. Our dialog and text completion experiments show that entropy-aware decoding is competitive with other decoding methods on quality, and repetitiveness while being more contextual. We hypothesize that the mostly deterministic nature of entropy-aware decoding will also improve the factuality of the generation, an important problem that needs to be solved before the wide-scale deployment of large language models. We leave this analysis for future work.

\section{Acknowlegment}
This research was enabled in part by compute resources and support provided by Calcul Québec\footnote{https://www.calculquebec.ca}, Compute Canada\footnote{https://www.computecanada.ca} and Mila IDT team. Timothy J. O'Donnell, Doina Precup, and Jackie C.K. Cheung are supported by the Canada CIFAR AI Chair program. We would also like to thank Khimya Khetarpal, Marc Bellemare, Jules Gagnon-Marchand, Andrei Romascanu, Meng Cao, Sumana Basu, and Andre Cianflone for their valuable inputs and discussions, and Khimya Kheterpal and Ilia Kulikov for the feedback on this paper.

\bibliography{custom,anthology}
\bibliographystyle{icml2023}

\newpage
\appendix
\onecolumn

\section{Unsmoothed And Smoothed Stable Entropy Baseline and Stable Entropy Zone}
\begin{figure}[h!]
  \centering
  \includegraphics[width=0.8\columnwidth]{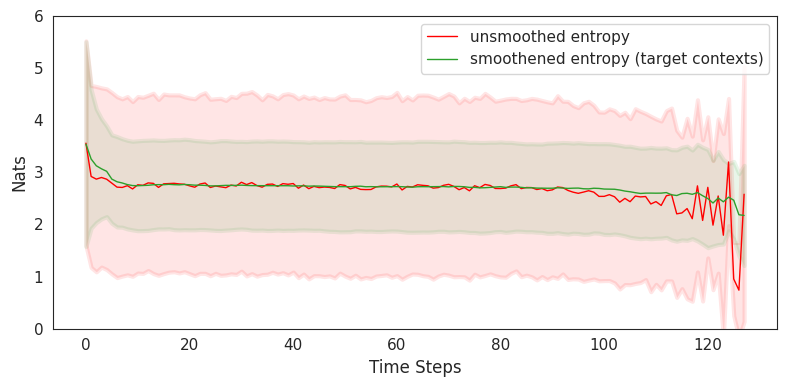}
  \caption{\textbf{Smoothed vs unsmoothed Stable Entropy Zone.} We plot the stable entropy baseline and the stable entropy zone using both the smoothed and unsmoothed entropy. In either case, we observe that the stable entropy baseline and the stable entropy zone remain mostly flat except for the first and last few time steps.}
  \label{fig:smoothed_vs_unsmoothed_stable_entropy_zone}
\end{figure}
\section{Additional Correlational Plots And Configurations for Stable Entropy Hypothesis Analysis}~\label{sec:additional_corr_plots}

\begin{figure*}[th!]
  \centering
  \begin{tabular}{@{}c@{}c@{}c@{}} 
    \begin{subfigure}{0.37\textwidth}
      \centering
      \includegraphics[width=1\textwidth]{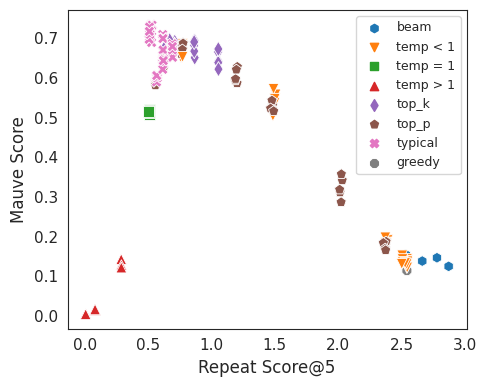}
      \caption{Repeat Score@5 vs Mauve Score} 
      \label{fig:repeat_vs_mauve}
    \end{subfigure} & 
    \begin{subfigure}{0.33\textwidth}
      \centering
      \includegraphics[width=1\textwidth]{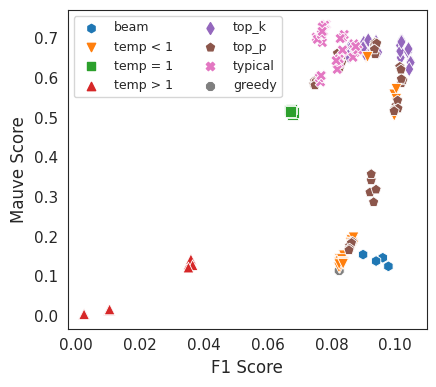}
      \caption{F1 Score vs Mauve Score} 
      \label{fig:f1_vs_mauve}
    \end{subfigure} &
    \begin{subfigure}{0.30\textwidth}
      \centering
      \includegraphics[width=1\textwidth]{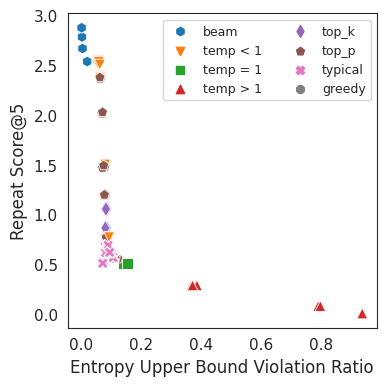}
      \caption{Repeat Score@5 vs EUVR.} 
      \label{fig:repeat_score_vs_euvr}
    \end{subfigure} 
  \end{tabular}
  \caption{\textbf{Additional Correlation Plots for Text Completion Experiments.} Figure (\protect\subref{fig:repeat_vs_mauve}) shows too many repeats (beam and greedy search, and temperature sampling ($T<<1$)) and too few repeats (for temperature sampling ($T>>1$)) both hurt generation quality. Figure (\protect\subref{fig:f1_vs_mauve}) shows that, among the stochastic decoding methods, top-k sampling balances the contextuality and generation quality conundrum the best. Finally, Figure~(\protect\subref{fig:repeat_score_vs_euvr}) shows a strong negative correlation between the repetition issue and entropy upper zone violations indicating that mostly lower-bound violations are mostly responsible for copying and repetitions.}
\label{fig:stable_entropy_hypothesis}
\end{figure*}
\clearpage
\section{Quantitative Results for Analysis of Various Decoding Algorithms for Text Completion Setup.}
\begin{table*}[th!]
  \small
  \centering  \ra{1.0}\scalebox{1}{
  \begin{tabular}{@{}p{0.28\textwidth}ccccccc@{}}\toprule
  \textbf{Sampling Method} & \textbf{F1} & \textbf{Rep. Score@5} & \textbf{3-gram rep.} & \textbf{Mauve} & \textbf{EVR} & \textbf{EUVR} & \textbf{ELVR} \\
  \midrule
  Greedy        &0.082 & 2.542 & 45.338 & 0.114 & 0.447 & 0.0560 & 0.391 \\
  \midrule
  Beam (n=5)    & 0.094 & 2.664 & 48.138 & 0.138 & 0.585 & 0.004 & 0.581 \\
  \quad +3-gram block & 0.102 & 0.666  &  0.063 & 0.476 & 0.170 & 0.014 & 0.155 \\

  \midrule
  \multicolumn{7}{@{}l}{Temperature Sampling} \\
  \midrule
  \quad $t=0.5$ & 0.100 & 1.499 & 16.159 & 0.537 & 0.238 & 0.078 & 0.160 \\
  \quad $t=0.8$ & 0.091 & 0.761 &  3.146 & 0.653 & 0.162 & 0.093 & 0.069 \\
  \quad $t=1$   & 0.068 & 0.511 &  1.015 & 0.507 & 0.193 & 0.155 & 0.038 \\
  \quad $t=1.2$ & 0.035 & 0.287 &  0.178 & 0.130 & 0.403 & 0.383 & 0.020 \\
  \midrule
  \multicolumn{7}{@{}l}{Top-k Sampling} \\
  \midrule
  \quad $k=30$  & 0.094 & 0.709 &  2.416 & 0.665 & 0.148 & 0.083 & 0.065 \\
  \quad $k=50$ & 0.091 & 0.666 & 2.016 & 0.667 & 0.144 & 0.083 & 0.062 \\
  \midrule
  \multicolumn{7}{@{}l}{Nucleus Sampling} \\
  \midrule
  \quad $p=0.95$ & 0.075 & 0.557 &  1.289 & 0.592 & 0.169 & 0.122 & 0.047 \\
  \quad $p=0.9$  & 0.082 & 0.620 &  1.701 & 0.642 & 0.150 & 0.094 & 0.056 \\
  \midrule
  \multicolumn{7}{@{}l}{Typical Sampling} \\
  \midrule
  \quad $\tau=0.2$ & 0.076 & 0.507 &  0.819 & 0.697 & 0.129 & 0.074 & 0.054 \\
  \quad $\tau=0.9$ & 0.082 & 0.615 &  1.725 & 0.622 & 0.154 & 0.093 & 0.061 \\
   \midrule
    Target completions & 1.000 & 0.605 & 1.381 & 1.000 & 0.136 & 0.0631 & 0.0731 \\
  \bottomrule
  \end{tabular}}
  \caption{\textbf{Quantitiative results for text completion analysis.} F1 score between the target and generated completion measures the contextuality of the generations. 3-gram repeats measure the extent of repetition problem with the generations. Entropy Lower-Bound Violation Ratio (ELVR), Entropy Upper-Bound Violation Ratio (EUVR), and Entropy Violation Ratio (EVR) measure the frequency with which entropy lower-bound, entropy upper-bound, and both combined are violated.}
  \label{tab:stable_entropy_quantitative_results}
\end{table*}
\clearpage
\section{Mean Conditional Entropies of Various Decoding Algorithms}~\label{sec:mean_entropies_wiki}
\begin{figure}[h!] 
  \centering
    \includegraphics[width=0.5\columnwidth]{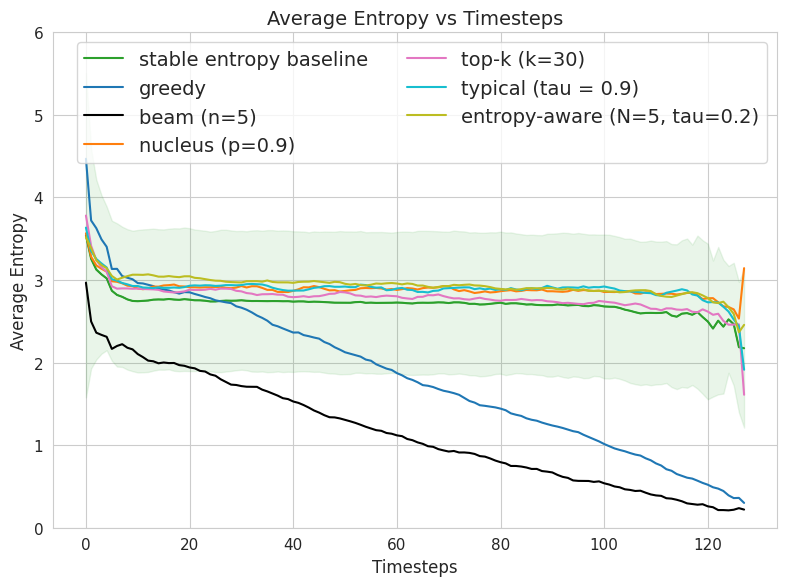}
  \caption{\textbf{Mean conditional entropies of various decoding algorithms.} We observe that the mean entropy under greedy and beam search drops near-monotonically over the sequence length. Well-tuned Sampling-based methods and entropy-aware decoding nearly follow the stable entropy baseline.}
  \label{fig:mean_conditional_entropies_wiki}
\end{figure}

\section{Qualitative Examples}
\subsection{Wikipedia Text Completion}
\begin{table*}[h!]
  \centering  \ra{1.4}
    \begin{tabular}{@{}p{1\textwidth}@{}}
      \textbf{Context}: ...With the coming of the Interstate Highway System in Michigan, the MSHD planned to convert the eastern section of US 2 to a freeway between St. Ignace and Sault Ste. Marie. In planning maps from 1947, this highway corridor was included in the system that later became the Interstates.\\ \hline\hline
      \textbf{Greedy}: The western \textcolor{red}{section of US 2 was renumbered as} M‑54 in 1949. The \textcolor{red}{eastern section of US 2 was renumbered} as M‑28 in 1950. \textcolor{red}{The western section of US 2 was renumbered as M‑54 in} 1951. \textcolor{red}{The eastern section of US 2 was renumbered as M‑28 in} 1952. \textcolor{red}{...}\\ \hline
      \textbf{BeamSearch} (k=5): In the 1950s and 1960s, the \textcolor{red}{M‑54 designation was} used to designate the \textcolor{red}{section of US 2} between \textcolor{red}{Sault Ste. Marie and Sault Ste. Marie. In the} 1970s, \textcolor{red}{the M‑54 designation was used to designate the section of US 2 between Sault Ste. Marie and Sault Ste. Marie. In the} 1980s, \textcolor{red}{the M‑54 ...}\\ \hline
      \textbf{Top-k Sampling} (k=30): In 1948, the western end of US 2 was rerouted to the eastern point near the old US 2/US 3 junction in Sault Ste. Marie. A section of the original M‑1 was restored to US 2 at Wakefield before the first interstate was completed. ...\\ \hline
      \textbf{Typical Sampling} ($\tau$=0.9): Michigan Department of Transportation letters to Congress, dated 18 November 1947 and 3 June 1948, approved the abandonment of the business US 2 segment of US 2 in favor of an expanded US 12. Approximately 3.5 miles of US 12 was completed between Ann Arbor and Sault Ste. Marie between 1964 and 1965. ...\\ \hline
      \textbf{Entropy-Aware Decoding}: ... The western section of US 2 was renumbered as M‑47 in December 1949. With the interchange in Epoufette opened in 1949, the state route continued to serve businesses in M‑48. A traffic jam of 3,000 cars per hour formed on US 2 in July 1950 when three women got into a disagreement on the roadway. After about an hour, the cars were cleared and the issue resolved. ... \\
      \bottomrule 
    \end{tabular}
    \caption{Generation examples using various decoding methods in a text completion setting using GPT-2 XL model. Greedy and beam search results in catastrophic degeneration (repetitions highlighted in red) whereas stochastic methods generate relatively more coherent completions.}
    \label{tab:decoding_generations}
\end{table*}
\clearpage
\subsection{Dialog Generation}
\begin{table*}[h!]  
\centering \ra{1.1}
\begin{tabular}{@{}ll@{}}
  \toprule
  \ctxt{
    \textbf{Person 1:} I do not like the taste, texture or thought of eating meat.} \\
  \ctxt{\textbf{Person 2:} So it has nothing to do with whether or not eating meat is ethical? There is a lot of debate around the ethics of eating meat.
  } \\
  \ctxt{\textbf{Person 1:}} \vspace*{0.5em} \\
  \bb{Beam Search} & I don't \textcolor{red}{like the taste,} texture, \textcolor{red}{or thought of eating meat. I don't like the taste, texture, or thought of eating meat.} \\ 
  \hdashline 
  \bb{EAD} & It's always good to hear different opinions. What would you say is the reason you don't eat meat? \\
  \midrule
  \ctxt{\textbf{Person 1:} Sounds like you had a great childhood then!}\\
  \ctxt{\textbf{Person 2:} You could say that. What was your childhood like?}\\
  \ctxt{\textbf{Person 1:} I can't really remember a lot, but I did play a lot of basketball with my friends}\\
  \ctxt{\textbf{Person 2:} I love basketball! I took part in many basketball tournaments and played in high school as well. What position do you play?}\\
  \ctxt{\textbf{Person 1:} I still practice basketball regularly, not a particular position thought}\\
  \ctxt{\textbf{Person 2:}}\vspace*{0.5em}\\
  \bb{Beam Search} & I played \textcolor{red}{a lot of basketball in high school as well}. I was a point guard. \textcolor{red}{What position do you} play? \\
  \hdashline
  \bb{EAD} & I still play as well. I really enjoy it. I'm good at it. I usually play with my friends. \\ 
  \midrule
  \ctxt{\textbf{Person 1:} We want to have at least 6 kids. We are from Italy, and we love having big families!} \\
  \ctxt{\textbf{Person 2:} Oh, that is awesome! Do you and your wife come from a big family?}\\
  \ctxt{\textbf{Person 1:} My wife does, I am an only child. My mom is in the medical field and did not really have time for kids.} \\
  \ctxt{\textbf{Person 2:} I'm an only child as well. I wish that I had siblings growing up but it did allow me special moments with my parents.}\\
  \ctxt{\textbf{Person 1:}}\vspace*{0.5em}\\
  \bb{Beam Search} & My wife and I \textcolor{red}{want to have at least 6 kids. We are from} Italy and \textcolor{red}{want to have} a big family.\\
  \hdashline
  \bb{EAD} & I always wanted siblings, but my parents did not want me to have any. \\
  \bottomrule
\end{tabular}
\caption{Dialog qualitative examples where beam search produces at least two 3-gram repeats.}
\label{tab:safe_examples}
\end{table*}

\subsection{Beam Search With Various Models in \autoref{tab:beam_search_entropy_violations}}~\label{fig:enc_dec_beam_mean_conditional_entropies}

\section{Models and Hyperparameters}~\label{sec:model_and_hyperparams}
\subsection{Various Configurations of Decoding Algorithm Evaluated in Section~\ref{sec:stable_entropy_experiments}}~\label{sec:decoding_configs}
We evaluate the following configurations of stochastic decoding algorithms for the stable entropy hypothesis experiments. We run each algorithm on three different seeds.
\begin{itemize}
  \item Top-K Sampling ($k$): 5, 10, 30, 50, 100,
  \item Nucleus Sampling ($p$): 0.15, 0.25, 0.4, 0.5, 0.75, 0.9, 0.95,
  \item Ancestral Sampling with Temperature ($t$): 0.001, 0.01, 0.1, 0.2, 0.5, 0.8, 1.0, 1.2, 1.5, 3.0,
  \item Typical Sampling ($\tau$): 0.2, 0.25, 0.5, 0.75, 0.9, 0.95.
\end{itemize}

\section*{State of the Art Models in \autoref{tab:beam_search_entropy_violations}}
\begin{itemize}
  \item CNN-DM: \citet{zhao2022calibrating}
  \item ArXiv: \citet{pang2022long}
  \item WMT: \citet{ng-etal-2019-facebook}
\end{itemize}
\section{Repeat Score@5}~\label{sec:repeat_score_5_def}
We modify the Repeat Score@5 metric proposed in ~\cite{aroraDIRECTORGeneratorClassifiersSupervised2022}, to capture the repetition at various n-gram levels. The modified metric is just a length-normalized version of the original metric. We compute Repeat Score@5 as
\begin{align}
  \text{Repeat Score@5} = \log_{2} \Bigg(\frac{\sum_{i=1}^{5} 2^{i} \times \text{\# i-grams}}{\text{\# cuml n-grams}} \Bigg) \times \text{\# 1-grams}/\# Tokens
\end{align}

where $\text{\# cuml n-grams} = \sum_{i=1}^{5} \text{\# i-grams}$. 

Intuitively, the metric captures average numbers of repetitions per token while exponentially weighing 1-gram to 5-gram repetitions.

\section{Entropy Aware Decoding Analysis}
\begin{figure*}[h!]
  \centering
  \begin{tabular}{@{}c@{}c@{}c@{}} 
    \begin{subfigure}{0.33\textwidth}
      \centering
      \includegraphics[width=1\textwidth]{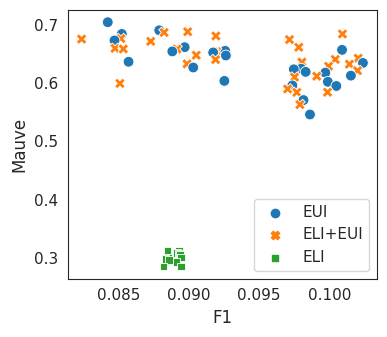}
      \caption{ELI vs EUI vs ELI+EUI} 
      \label{fig:ead_eli_vs_eui_vs_ei}
    \end{subfigure} & 
    \begin{subfigure}{0.33\textwidth}
      \centering
      \includegraphics[width=1\textwidth]{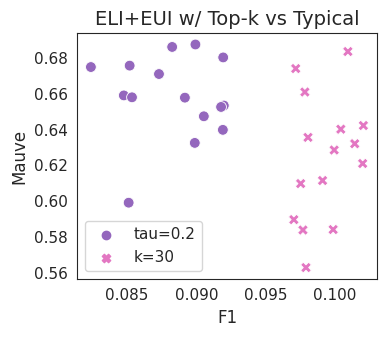}
      \caption{Top-k vs Typical Sampling} 
      \label{fig:ead_topk_vs_typical}
    \end{subfigure} &
    \begin{subfigure}{0.33\textwidth}
      \centering
      \includegraphics[width=1\textwidth]{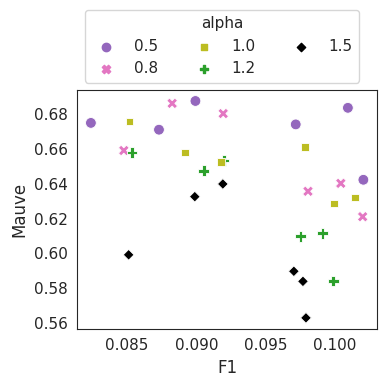}
      \caption{Repeat Score@5 vs EUVR.} 
      \label{fig:ead_std_dev}
    \end{subfigure} 
  \end{tabular}
  \caption{\textbf{Entropy-Aware Decoding Analysis.} Figure (\protect\subref{fig:ead_eli_vs_eui_vs_ei}) shows that only entropy lower-bound interventions (ELI) do not work and most gains come from entropy upper-bound interventions. Figure (\protect\subref{fig:ead_topk_vs_typical}) shows that top-k sampling with EAD results in a higher F1 score. Finally, Figure~(\protect\subref{fig:ead_std_dev}) shows that high $alpha$ values result in poorer performance, hence it is reasonable to use the band of $\le 1$.}
\end{figure*}

\clearpage
\section{Surprisal Under Different Decoding Algorithms}
\begin{figure*}[h!] 
  \centering
    \includegraphics[width=1\textwidth]{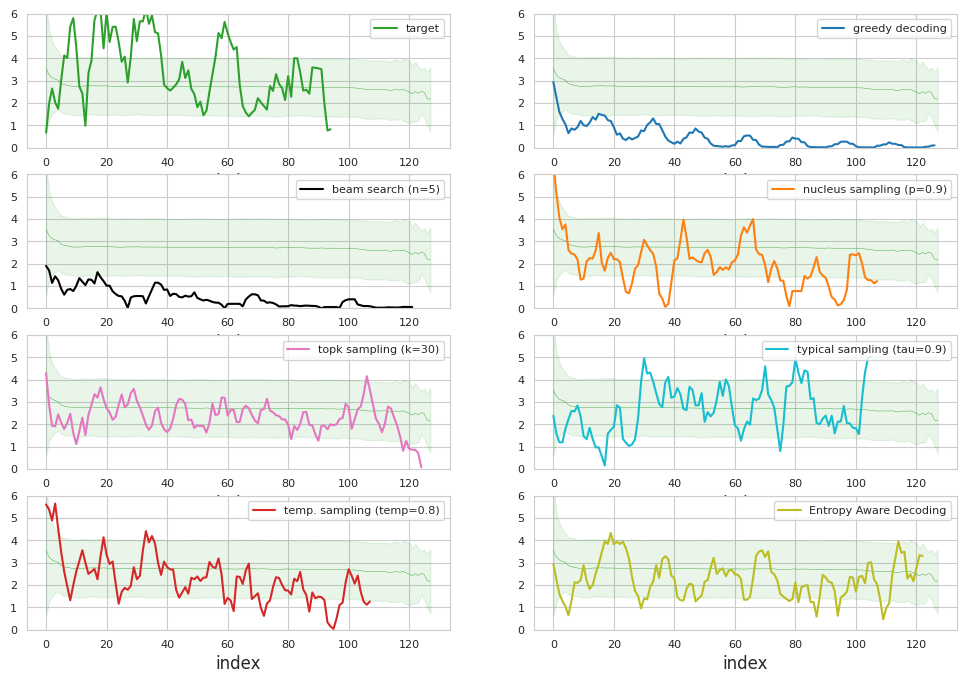}
  \caption{\textbf{Visualization of surprisal of various decoding algorithms.} Visualizing the smoothed surprisal (smoothing window size $5$) for various decoding algorithms in a text completion setup for the prompt from \autoref{tab:decoding_generations}. The faint green line in the background is the stable entropy baseline and is used to represent the target information rate. We observe the catastrophic drop in surprisal for beam and greedy search. Stochastic algorithms oscillate near the target information rate.}
  \label{fig:decoding_algorithm_surprisal_visualizations}
\end{figure*}
In the figure, we also find that stochastic decoding methods and entropy-aware decoding induce a context distribution that results in surprisal centered around the target information rate whereas the surprisal under greedy and beam search decoding suffer from a similar catastrophic drop as observed in \autoref{fig:decoding_algorithm_entropy_visualizations}.

\end{document}